\documentclass[cameraready]{Interspeech}

\usepackage{graphicx}
\usepackage{booktabs}
\usepackage{array}
\usepackage{multirow}
\usepackage{url}

\title{Shifting Relational Paradigms for Affective Computing: Affective Resonance, Vitality Affects, and Vocal Interaction Fields}
\author[affiliation={1}, orcid=0009-0005-2149-7372, correspondingauthor]{Cy}{Gorman}
\author[affiliation={1}]{Yihang}{Yao}
\address{
    $^1$ Nurobodi, Australia
}
\email{info@nurobodi.com}

\begin{document}
\maketitle

\begin{abstract}
Affective computing has largely followed an individual-state paradigm, extracting discrete emotion labels or arousal/valence from isolated speakers. We argue this framing is incomplete for interaction. Drawing on affective resonance and vitality-contour accounts, we propose a relational framework in which the primary unit of affective analysis is the interactional field constituted within vocal dynamics. As a proof of concept, we present a preliminary empirical study using continuous self-supervised speech representations to detect directional expressive coupling in multi-party conversation. Coupling is regime-specific, concentrated at sub-second timescales, and collapses under exclusive-speech negative controls, consistent with a relational account of affective dynamics. We introduce design frameworks for Artificial Affective Resonance Intelligence grounded in Affective Resonance Dynamic Ontologies, supported by null-calibrated directional coupling analyses across interaction regimes.
\end{abstract}

\section{Introduction}
Affective computing --- the endeavour of building systems that recognise, model, and respond to human emotion --- has largely proceeded by asking what emotion a person is likely experiencing within a given interactional context. Dominant pipelines map acoustic, linguistic, physiological, or biometric signals to discrete emotion categories or continuous arousal/valence dimensions \cite{ref01,ref02}. This paradigm remains productive, but recent advances in cybernetic architectures, multimodal sensing, and interaction design open a complementary modelling focus: treating relational dynamics as a primary site of affective organisation. The relational field is not restricted to any single expressive modality; here, we examine it through vocal dynamics as a tractable and theoretically motivated domain. As Kappas and Gratch \cite{ref03} observe, affective computing lacks robust models of how affective behaviour unfolds relationally over time. M{\"u}hlhoff \cite{ref04} proposes that affective resonance is not the outcome of individuals converging toward similar internal states, but an emergent property of the interactional field they co-constitute ---  a dynamic individual-convergence models cannot capture because they treat affect as an attribute of persons rather than a property of interaction. Acoustic entrainment research demonstrates convergence in vocal features over time \cite{ref05,ref06,ref07}, including via specific interactional events such as backchannels \cite{ref08}, but typically operationalises coupling as growing feature similarity between speaker trajectories rather than directionality, interactional configuration, or field-level emergence. Regime-dependent coupling that collapses under exclusive speech would indicate that interactional structure, not individual state, governs expressive dynamics --- a pattern existing frameworks cannot detect. Granger-causal approaches to speech coupling remain comparatively rare and are typically bivariate and single-feature \cite{ref09,ref10}, with no prior work, to our knowledge, combining conditional multi-speaker models with interaction-regime decomposition. Emotion Recognition in Conversations extends affective computing to multi-party settings but aggregates individual-state inferences rather than modelling speakers' relational dynamic \cite{ref03}. Pre-semantic vocal resonance across interacting speakers remains underexplored as a computational target despite its role as a direct marker of vitality dynamics \cite{ref11,ref12}. We propose a relational reorientation toward field-level models of human interaction, grounded in M{\"u}hlhoff's account of affective resonance \cite{ref04}, Stern's theory of vitality affects and contours \cite{ref11}, as extended by K{\o}ppe et al \cite{ref12}, and enactivist accounts of participatory sense-making \cite{ref13,ref14}. We ground this position theoretically, demonstrate it through directional coupling analysis of vocal interaction, and develop design implications --- Affective Resonance Dynamic Ontologies (ARDO) and Artificial Affective Resonance Intelligence (AARI) --- for systems oriented toward interactional dynamics rather than individual state inference. Our empirical contribution comprises: (1) a null-calibrated directional coupling framework correcting for size inflation in short-window autoregressive testing; (2) an interaction-regime decomposition enabling within-corpus empirical controls; (3) a multi-speaker conditional analysis reducing spurious dyadic coupling from group-level confounds; and (4) results on the AMI Meeting Corpus showing coupling concentrated at sub-second timescales under mutual co-presence, while macro-scale effects are provisional. This work directly addresses the Interspeech 2026 theme, Speaking Together, by treating speech as interactional coordination rather than individual production.

\section{Theoretical foundations}
\subsection{Affective resonance as emergent field property}
M{\"u}hlhoff \cite{ref04} proposes that affective resonance names a field of phenomena in which interactional dynamic flux constitutes affective experience, rather than transmitting pre-existing inner states between independently existing individuals. He identifies three constitutive axioms: (1) affective resonance is a dynamical entanglement of moving and being-moved in relation; (2) it is primarily experienced as an immanent force arising within relational interplay; and (3) it is an emergent, creative dynamic, producing its own vectors of movement-in-relation, rather than operating within pre-formed affective state spaces defined by pre-categorised emotion labels. This reflects an ontological commitment distinct from convergence-based paradigms where entrainment research typically operationalises coupling as convergence in feature values \cite{ref05,ref06}, treating interaction as alignment between individual trajectories rather than as a field-level dynamic with its own organisation. By contrast, M{\"u}hlhoff's account gives ontological primacy to the relational dynamic --- subjectivity and individual affective experience arising from life-long histories of being-in-relation --- rather than to pre-formed individuals who subsequently interact. Affective resonance is therefore distinct from imitation, synchrony, or mimicry; it does not require congruence of form, but the immanent connectivity of forces that constitutes the field as an emergent property of interaction. Understanding this temporal-dynamic texture requires a finer-grained account of the expressive substrate through which resonance becomes perceptible.

\subsection{Vitality affects, vitality contours and pre-semantic voice}
Stern \cite{ref11} introduced vitality affects as dynamic, non-categorical qualities and later described vitality contours---surging, fading, rushing, bursting---as their temporal form. K{\o}ppe, Harder, and V{\ae}ver \cite{ref12} clarify three central dimensions: temporality, intensity, and form. These dynamics are amodal and cross-modal; here we focus on voice as both a theoretical and practical carrier. Vocalics (e.g., timbre, formants, harmonic ratios, jitter, shimmer) form a continuous energetic substrate, while prosody organises it through intonation, stress, rhythm, and pitch contour. We use \emph{prosodic} broadly to emphasise the pre-semantic organisation of temporal-affective contours, potentially underpinned by microtonal and resonant features. Their continuous co-modulation traces the ``temporal feeling shapes'' through which vitality becomes interactionally available \cite{ref11}. This motivates retaining continuous hidden-state representations rather than discrete tokens. WavLM \cite{ref15}, HuBERT \cite{ref16}, and wav2vec 2.0 \cite{ref17} encode paralinguistic and prosodic information without task annotation, as demonstrated by SUPERB \cite{ref18} and speech-emotion studies \cite{ref19,ref20}. Continuous representations preserve temporal dynamics that clustering can suppress: discrete speech tokens tend toward phonetic invariance at the expense of prosodic correlation \cite{ref21,ref22,ref23}. This matters because vitality is upstream of categorical emotion. Emotional prosody elicits neural responses at roughly 200 ms \cite{ref24}, before semantic processing around 300--400 ms \cite{ref25}. Category-based prosody and emotion systems \cite{ref02,ref20,ref26} therefore model derived representations rather than the force, movement, directionality, and aliveness through which affective experience is organised. Computational access is most consequential at this pre-semantic level, where the interaction field's temporal, intensive, and formal modulations are directly expressed.

\subsection{Enactivism and the relational subject}
The enactivist tradition \cite{ref13} grounds cognition in meaning brought forth through embodied interaction rather than internal representation. De Jaegher and Di Paolo \cite{ref14} extend this through participatory sense-making: meaning is jointly constituted in interaction itself, not carried by pre-formed individuals. The relational subject emerges within interaction rather than prior to it --- a claim that directly shapes what it means to design a system that participates in an interactional field rather than observes one.

\subsection{ARDO and AARI: interaction field modelling}
These foundations motivate a working design distinction. Affective Resonance Dynamic Ontologies (ARDO) names interaction-level organisation that becomes salient when sustained vocal dynamics are modelled as a field rather than as independent states. Following M{\"u}hlhoff \cite{ref04}, discrete states attributed to people or an AI are downstream products of resonance, not its preconditions. This commitment shapes feature selection, training objectives, and outputs, and makes ARDO a frame for testing whether interaction dynamics contain organisation irreducible to individual trajectories. Artificial Affective Resonance Intelligence (AARI) names systems built within that frame: they map the field's vitality trajectories and generate responses calibrated to its modulation rather than classify individual affect. Conventional categorical or dimensional systems \cite{ref02} expose confidence scores over fixed labels; useful as interfaces, these presuppose the categories whose emergence a relational account seeks to explain. ARDO instead targets continuous tension, volatility, and momentum that may later organise into recognisable emotions; AARI is the corresponding architectural programme.

\section{Empirical illustration}
\subsection{Rationale}
The framework predicts that directional expressive coupling should appear when mutually oriented speakers constitute an interactional field and weaken when that condition is removed. Tier A captures simultaneous co-presence; Tier B pools non-overlapping single-speaker activity, which may include sequential engagement \cite{ref14}; Tier C separates that activity by speaker direction. We test this through interaction-regime decomposition, providing within-corpus empirical controls. The study is a conceptual proof of principle; complete implementation details are reserved for subsequent work.

\subsection{Method}
\noindent\textbf{Data:} We analyse four-speaker meeting audio from the AMI Meeting Corpus \cite{ref27} using close-talking headset channels, resampled to 16 kHz and processed in non-overlapping 60-second windows. Voice activity detection is applied per speaker and mapped to the WavLM feature-frame grid; the effective frame hop is determined by the model's temporal downsampling ($\approx$20 ms at 16 kHz), and we align masks by window-length-to-T mapping to ensure exact synchrony; all results reported here are AMI-only, with generalisation to other corpora reserved for future work.

\noindent\textbf{Feature extraction:} We extract continuous hidden-state representations from the 12 transformer layers of WavLM-Base+ \cite{ref15} --- a self-supervised speech model pre-trained on large-scale audio. (The HuggingFace API also returns an initial embedding state; in our notation we use \texttt{hidden\_states[1..12]} as the transformer layers.) Crucially, we do not apply quantisation or K-means clustering, which discard prosodic and affective information in favour of phonetic invariance \cite{ref21,ref22}. We retain continuous floating-point hidden states, which preserve the dense entanglement of acoustic, prosodic, and paralinguistic information across representational levels --- the representational substrate ARDO requires for tracking vitality dynamics. Our primary feature --- expressiveness --- is a cross-layer activation dispersion statistic derived from WavLM hidden-state dynamics, intended to capture moment-to-moment richness and complexity of vocal expression beyond overall magnitude.

\noindent\textbf{Multi-dimensional proxies and energy control:} Beyond a scalar energy proxy, the same representation pass yields aligned series for cross-representational expressiveness, high-level semantic magnitude, and frame-to-frame topic drift. All share the same feature-frame axis, windowing, and regime decomposition. A regime-contiguous episode is a maximal uninterrupted sequence of frames within a window for which a selected pre-specified dyadic VAD mask remains true. To test whether expressiveness merely shadows the energy proxy, we residualise it against energy within each regime-contiguous episode before directional testing and null calibration. This targets temporal-dynamic expressive richness, consistent with vitality contours (Sec. 2.2), rather than categorical affect.

\noindent\textbf{Interaction condition decomposition:} For each speaker pair within each window, we define three conditions via binary voice activity masks, with minor temporal smoothing applied to bridge brief within-speaker gaps. Tier A denotes overlapping speech. Tier B pools non-overlapping single-speaker activity across both active-speaker directions. Tier C separates that non-overlapping activity according to which speaker is active. These conditions compare coupling across overlapping, pooled non-overlapping, and direction-specific non-overlapping activity. Tier C provides an AMI-specific empirical control by separating non-overlapping activity according to which speaker is active.

\begin{figure}[!t]
    \centering
    \includegraphics[width=\columnwidth]{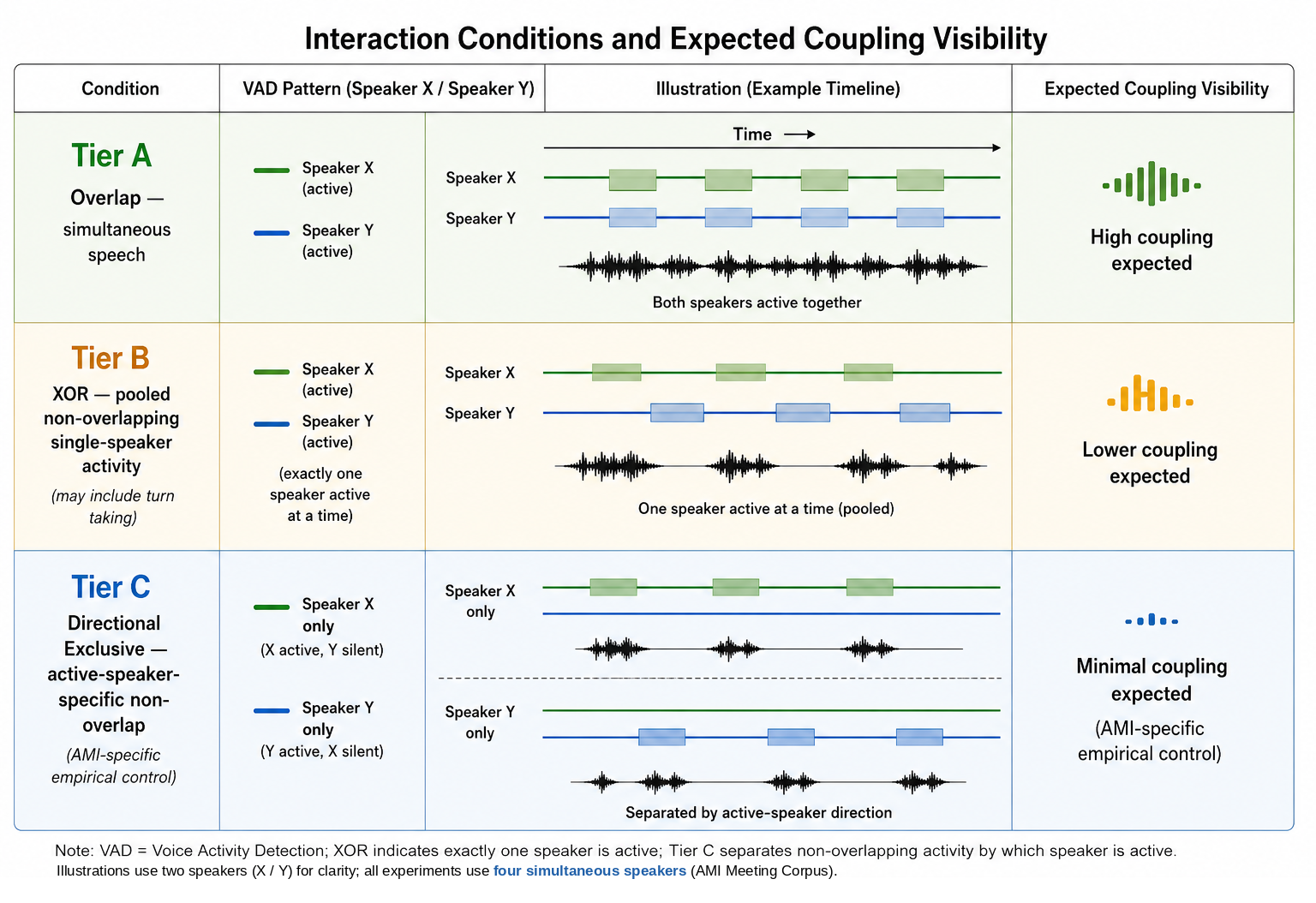}
    \caption{Interaction conditions and expected coupling visibility. Tier A denotes overlapping speech; Tier B pools non-overlapping single-speaker activity; Tier C separates that activity by active-speaker direction and is used as an AMI-specific empirical control.}
    \label{fig:regimes}
\end{figure}

\noindent\textbf{Causal analysis:}We test directional Granger causality \cite{ref28} using bivariate and 4-speaker conditional vector autoregression (VAR) models \cite{ref10,ref28}. In the conditional model, each dyadic test conditions on the other two speakers to reduce group-level confounds such as shared laughter, room events, and topic shifts. Granger tests are used here to assess linear predictive dependence within a VAR framework, not to claim that affective interaction dynamics are themselves linear; nonlinear extensions such as transfer entropy remain a future direction. Within 60 s windows, VAR/Granger is applied to regime-contiguous episodes; their short duration makes lag-selection size inflation a central concern. We therefore compare observed rejection behaviour with matched circular-shift nulls and report null-quantile calibrated excess rejections (OBS--NULL, percentage points; Fig. 2). Here, $n_{\mathrm{obs}}$ counts episode-level tests and can exceed the number of contributing windows; negative values mean fewer rejections than the calibrated null. Window-level directionality is summarised by a signed Directionality Support Score (DSS $\in \{-1,0,+1\}$) after false-discovery-rate control \cite{ref29}, using both Window-BH and conservative Global-FDR. We call lags $\leq 1$ s micro and those $>1$ s macro; exact lag grids and gates are implementation-specific.

\begin{table}[!t]
\caption{Directionality Support Score (DSS) distribution (\% of windows with DSS $=-1,0,+1$) under within-condition correction (Window-BH) and conservative global correction (Global-FDR), for micro- and macro-scale lags. $n$ denotes the number of evaluated windows per tier.}
\label{tab:dss}
\centering
\scriptsize
\setlength{\tabcolsep}{1.5pt}
\renewcommand{\arraystretch}{0.92}
\begin{tabular}{llrrrrrrr}
\toprule
Scale & Tier & $n$ & WB$-1$ & WB0 & WB$+1$ & GF$-1$ & GF0 & GF$+1$ \\
\midrule
micro & A & 502 & 23.7 & 52.4 & 23.9 & 17.3 & 61.0 & 21.7 \\
micro & B & 71  & 14.1 & 66.2 & 19.7 & 9.9  & 88.7 & 1.4 \\
micro & C & 62  & 0.0  & 98.4 & 1.6  & 3.2  & 95.2 & 1.6 \\
macro & A & 282 & 14.5 & 74.1 & 11.3 & 2.1  & 95.7 & 2.1 \\
macro & B & 18  & 22.2 & 77.8 & 0.0  & 22.2 & 72.2 & 5.6 \\
macro & C & 87  & 12.6 & 71.3 & 16.1 & 0.0  & 95.4 & 4.6 \\
\bottomrule
\end{tabular}
\end{table}

\subsection{Results}
Directional coupling was reliably detectable at sub-second lags under interaction regimes involving mutual orientation. Window-level directionality is sparse but structured: micro-scale overlap (Tier A) shows the highest non-zero support mass, with directional wins that are near-symmetric across directions (Table 1). Tier B (pooled non-overlapping activity) and Tier C (directional exclusive activity) shift strongly toward zero support, consistent with weaker or absent dyadic coupling outside mutual co-presence. Matched circular-shift results confirm short-series size inflation (Table 2), motivating calibrated excess rejection rates (Fig. 2). At $q=5\%$, Tier A micro shows approximately +6.6 percentage points under the bivariate model and +7.4 points under energy-controlled conditional analysis. At the stricter $q=1\%$ operating point (not shown), effects attenuate but remain regime-specific: Tier A is strongest, Tier B weaker, and Tier C approaches the null. Coupling therefore cannot be reduced to synchrony in the energy proxy alone and is concentrated at short temporal offsets relevant to an AARI system. Non-zero directionality is feature-selective across regime and timescale; small but structured cross-feature effects, tested bivariately, support a multidimensional account of vitality dynamics beyond a single scalar energy mechanism

\begin{table}[!t]
\caption{Shift-null rejection rates under nominal thresholds for the full shift-null grid ($n_{\mathrm{null}}$ shown), illustrating size inflation in short-series lag selection settings and motivating null calibration.}
\label{tab:null}
\centering
\scriptsize
\setlength{\tabcolsep}{1.8pt}
\renewcommand{\arraystretch}{0.92}
\begin{tabular}{llrrrrr}
\toprule
Tier/scale & Mode & $n_{\rm null}$ & NULL $p<.05$ & NULL $p<.01$ & Infl.$<.05$ & Infl.$<.01$ \\
\midrule
A/micro & overlap & 20745 & 11.9\% & 5.9\% & 2.4$\times$ & 5.9$\times$ \\
A/macro & overlap & 2780  & 17.4\% & 12.8\% & 3.5$\times$ & 12.8$\times$ \\
B/micro & xor     & 8545  & 9.9\%  & 3.5\% & 2.0$\times$ & 3.5$\times$ \\
B/macro & xor     & 820   & 12.6\% & 5.4\% & 2.5$\times$ & 5.4$\times$ \\
C/micro & x\_not\_y & 2460 & 10.4\% & 3.7\% & 2.1$\times$ & 3.7$\times$ \\
C/micro & y\_not\_x & 3380 & 10.7\% & 3.6\% & 2.1$\times$ & 3.6$\times$ \\
C/macro & x\_not\_y & 460  & 10.0\% & 3.7\% & 2.0$\times$ & 3.7$\times$ \\
C/macro & y\_not\_x & 760  & 8.7\%  & 2.5\% & 1.7$\times$ & 2.5$\times$ \\
\bottomrule
\end{tabular}
\end{table}

\begin{figure}[!t]
    \centering
    \includegraphics[width=\columnwidth]{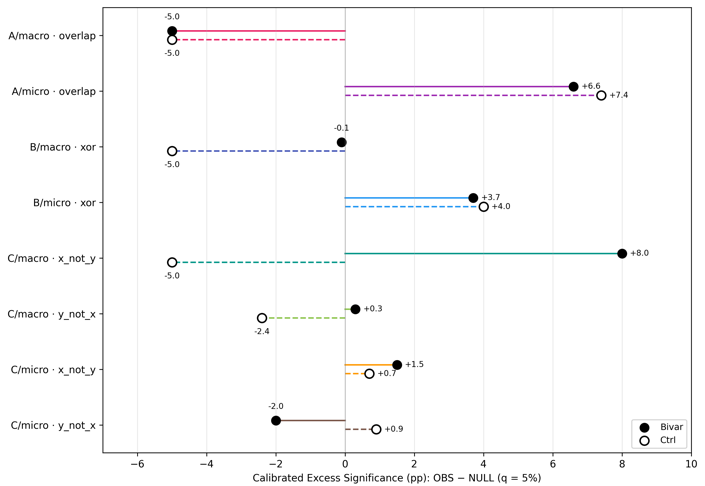}
    \caption{Calibrated excess significance (OBS $-$ NULL, $q=5\%$, percentage points) across representative operating points. Filled markers denote bivariate tests; open markers denote 4-speaker control; values $<0$ indicate fewer rejections than the null baseline.}
    \label{fig:calibrated}
\end{figure}

\subsection{Limitations} The expressiveness proxy measures modulation in latent vocal-production dynamics, not valence or discrete emotion; this is deliberate. Granger analysis captures linear predictive dependence but does not distinguish convergent from divergent coupling, for which VAR coefficient signs are a principled next step. Results are AMI-only, limiting immediate generalisation, although they establish regime- and scale-conditioned coupling within formal four-speaker AMI meeting interaction. Cross-corpus analysis has been completed and full results will be reported in subsequent work. Tier B macro analysis is underpowered (n=18 gated windows), and the DSS depends on FDR and minimum run-fraction parameters; conservative Global-FDR provides only partial robustness. Future work should expand sensitivity analysis, corpora, and features targeting temporality, intensity, and form \cite{ref12}, including multimodal extensions warranted by the amodal scope of vitality affects.

\section{Design implications}
The empirical results described above demonstrate that the interactional field leaves measurable traces: coupling is constituted by conversational configuration rather than carried into interaction by individual speakers --- a finding that individual-state pipelines are structurally unable to exploit. This opens the design question of how to build systems responsive to the interactional field as such, rather than to individual trajectories that only approximate it. This is where ARDO and AARI move from theoretical proposals to concrete design constraints. A system designed to participate in a resonance field rather than observe it from outside must make fundamentally different computational choices. Conventional affective computing pipelines infer individual affective states and respond to those inferences. By contrast, an AARI system is situated within the interactional field: its outputs function as contributions to the evolving relational dynamic rather than as responses to a diagnosis. This reorientation has consequences for training objectives. Rather than minimising classification error against discrete emotion labels, systems oriented by ARDO are evaluated in terms of the quality of relational coupling they sustain--- whether they maintain asymmetric, responsive attunement to the user's affective tempo, analogous to the dynamics described in Stern's infant--caregiver model \cite{ref11}. Emotion labels remain useful as diagnostic and benchmark interfaces but are not the primary object of ARDO. The vocalic layer is where this participation becomes concrete. Treating vocalics as a design medium rather than solely as a diagnostic signal opens possibilities that categorical-output systems foreclose. Juslin and Laukka \cite{ref30} show that vocal expression and musical performance share acoustic cue patterns for communicating emotion, suggesting that expressive dynamics in vocal interaction may be analysable using formal tools developed for musical structure. Concepts such as tension and release, consonance and dissonance can, in principle, be applied to moment-to-moment contours of multi-party vocal interaction. The human voice exhibits a harmonic series and shifts in the distribution of energy across that series---and in relation to another voice---carry affect-relevant information that categorical labels compress. We treat this as a working hypothesis consistent with evidence that affective processing precedes semantic integration in auditory cortex \cite{ref24,ref25}. The PIAT/ARTIST model \cite{ref31} provides a precedent for architectures that internalise tonal structure from unsupervised exposure, suggesting self-supervised learning of vitality contour structure without categorical emotion supervision as a plausible design direction. Extended toward generative systems, this points toward Human-AI cybernetic loops in which a synthetic vocal output participates in, rather than merely observes, the shared resonance field. ARDO formalises, at the level of design, implications already articulated across M{\"u}hlhoff's relational ontology \cite{ref04}, enactivist participatory sense-making \cite{ref14}, and Stern's developmental account of affective attunement \cite{ref11}. What ARDO adds is an evaluative standard: whether a computational approach models upstream resonance dynamics or only their downstream categorical interfaces. AARI foregrounds this distinction architecturally. Compassion-focused technology frameworks \cite{ref32} and relational ethics accounts of data-centric systems \cite{ref33} converge on the same insight: that systems operating within human affective life require attunement as a design condition, not an ethical afterthought. An AARI system instantiates this orientation at the level of design: responding to vitality dynamics within a resonance field rather than extracting affective signals for classificatory or commercial purposes. Entrainment research models convergence between individually defined speakers \cite{ref05,ref06,ref07,ref08}; AARI treats the interaction field as the primary object of modelling, with individual trajectories as observable traces of field-level dynamics.

\section{Conclusions}
Affective computing requires a reorientation from modelling individual affective states to modelling interactional fields as primary sites of affective organisation. M{\"u}hlhoff's account of affective resonance \cite{ref04}, Stern and K{\o}ppe's theory of vitality affects \cite{ref11,ref12}, and enactivist approaches to participatory sense-making \cite{ref13,ref14} provide the philosophical substrate for this shift. Our Empirical Illustration supports this reframing by showing that vocal coupling is regime-conditioned and temporally localised, collapsing when mutual orientation is removed --- consistent with a relational account of affective resonance rather than individual-state convergence. ARDO and AARI, introduced above, articulate the design implications of this reorientation: treating the interaction field, rather than individual states, as the primary object of modelling, and the vocalic layer as a site of expressive responsiveness, not a diagnostic channel.

\clearpage

\section{Acknowledgments}
The authors thank CSIRO, including the Innovate to Grow and ON Prime programme teams, for their support of Nurobodi's research and development, and ongoing research translation for industry.  We also thank Dr.\ Haytham Fayek for his valuable guidance and feedback, and for his ongoing encouragement to pursue this area of affective computing research.

\section{Generative AI Use Disclosure}
Generative AI tools (including large language models) were used to assist with editing, structural clarity, structural refinement, analysis and summarisation. AI tools were also used to support documentation of analysis workflows and preparation of visualisation scripts. All theoretical framing, experimental design and rationale, and implementation were produced by the authors, who retain full intellectual ownership and responsibility for the content of this work.

\end{document}